\documentclass[a4paper,11pt,twocolumn,twoside]{article}
\usepackage{sepln_en}
\usepackage{fullname}
\usepackage[utf8]{inputenc}
\input epsf

\usepackage{latexsym}

\usepackage{multirow}
\usepackage{xspace}
\usepackage{url}
\usepackage{tabularx}
\usepackage{microtype}
\usepackage{graphicx}
\usepackage{paralist}
\usepackage{subfig}
\usepackage{booktabs}
\usepackage{amsmath}
\usepackage{amssymb}
\usepackage{xstring}
\usepackage{scalefnt}
\usepackage{bm}
\usepackage{paralist}
\usepackage{tablefootnote}
\usepackage{subfig}
\usepackage{capt-of}
\usepackage{latexsym}
\usepackage{paralist}
\usepackage{calc}
\usepackage{booktabs}
\usepackage{scrextend}
\usepackage{color}
\usepackage{lmodern,textcomp}

\newcommand\footnoteref[1]{\protected@xdef\@thefnmark{\ref{#1}}\@footnotemark}

\definecolor{lightgreen}{RGB}{200,255,200}
\definecolor{lightblue}{RGB}{200,200,255}
\definecolor{lightred}{RGB}{255,200,200}
\definecolor{lighterred}{RGB}{255,220,220}

\newcommand{\R}{\mathbb{R}}

\newcommand{\original}{\textsc{Original}\xspace}
\newcommand{\mtreordered}{\textsc{Reordered}\xspace}
\newcommand{\nadj}{\textsc{Noun-Adj}\xspace}
\newcommand{\adjn}{\textsc{Adj-Noun}\xspace}
\newcommand{\random}{\textsc{Random}\xspace}
\newcommand{\onlylex}{\textsc{Only-Lexicon}\xspace}
\newcommand{\nolex}{\textsc{No-Lexicon}\xspace}

\newcommand{\bilstm}{\textsc{BiLstm}\xspace}
\newcommand{\cnn}{\textsc{Cnn}\xspace}

\newcommand{\bilstms}{\textsc{BiLstms}\xspace}
\newcommand{\cnns}{\textsc{Cnns}\xspace}
\newcommand{\Rnns}{\textsc{Rnns}\xspace}

\newcommand{\rt}[1]{\rotatebox{90}{#1}}

\newcommand{\ie}{\textit{i.\,e.}\xspace}
\newcommand{\eg}{\textit{e.\,g.}\xspace}
\newcommand{\F}{$\text{F}_1$\xspace}

\usepackage{annotates}

\usepackage{url}

\title{On the Effect of Word Order\\on Cross-lingual Sentiment Analysis}

\author {\textbf{Àlex R. Atrio$^1$}, \textbf{Toni Badia$^{2}$}, \textbf{Jeremy Barnes$^{3}$}\\[5pt]
$^1$ HEIG-VD \& EPFL \hspace{5pt}   $^2$Universitat Pompeu Fabra  \hspace{5pt}  $^3$University of Oslo\\
[1pt]{ \tt $^1$alejandro.ramirezatrio@heig-vd.ch \hspace{5pt} $^2$toni.badia@upf.edu \hspace{5pt} $^3$jeremycb@ifi.uio.no}
}

\seplntranstitle{Sobre el efecto del orden de las palabras\\en el análisis de sentimiento crosslingüe}

\seplnclave{análisis de sentimiento, crosslingüe, reordenamiento}

\seplnresumen{Los modelos de análisis de sentimiento que actualmente representan el estado del arte utilizan el orden
de las palabras, ya sea explícitamente al preentrenar con un objetivo de modelización del lenguaje, ya sea implícitamente al recurrir a redes neuronales recurrentes (RNR) o convolucionales (RNC). Esto es un problema para los acercamientos crosslingües que emplean vectores bilingües para entrenar, ya que la diferencia del orden de las palabras entre la lengua de origen y la de destino no se resuelve. En este trabajo, exploramos el reordenamiento de las palabras como etapa de procesamiento previa para la clasificación de sentimiento crosslingüe a nivel de frase, con dos combinaciones de idiomas (Inglés-Castellano, Inglés-Catalán). Descubrimos que aunque el reordenamiento ayuda a los dos modelos, los RNC son más sensibles al reordenamiento local, mientras un reordenamiento global beneficia a los RNR.}

\seplnkey{sentiment analysis, cross-lingual, reordering}

\seplnabstract{Current state-of-the-art models for sentiment analysis make use of word order
either explicitly by pre-training on a language modeling objective or implicitly
by using recurrent neural networks (\Rnns) or convolutional networks (\cnns).
This is a problem for cross-lingual models that
use bilingual embeddings as features, as the difference
in word order between source and target languages is
not resolved. In this work, we explore reordering
as a pre-processing step for sentence-level cross-lingual sentiment
classification with two language combinations
(English-Spanish, English-Catalan). We find that while reordering helps both models, \cnns are more sensitive to local reorderings, while global reordering benefits \Rnns.}

\firstpageno{1}

\begin{document}


\setlength\titlebox{16cm} 

\label{firstpage} \maketitle

%

\section{Introduction}

Cross-lingual Sentiment Analysis (CLSA) exploits resources, \eg labeled data of a high-resource language, to train a sentiment classifier for low-resource languages. This approach is useful when a target language lacks plentiful labeled data, particularly for specific domains. Machine Translation (MT) is often used to bridge the gap between languages \cite{Banea2008,Balahur2014d}, but requires abundant parallel data, which may be difficult to find for some low-resource languages. Approaches that use bilingual distributional representations, in contrast, have proven competitive while requiring less parallel data \cite{Chen2016,Barnes2018b}.

Recently, sentiment classifiers pre-trained on a language modeling task have lead to state-of-the-art results \cite{Peters2018,Howard2018,Devlin2018}. This improvement suggests that sentiment analysis benefits from learning word order and fine-grained relationships between tokens, which can be gleaned from unlabeled data. These approaches, however, have only been applied in a monolingual setting and it is not clear how the difference in word orders would affect them in a cross-lingual setup. In this work, we perform an analysis of the effect of word order on cross-lingual sentiment classifiers that use bilingual embeddings as features. We show that these models are sensitive to word order and benefit from pre-reordering the target-language test data so that it resembles the source-language word order.

\section{Related Work}

\paragraph{Cross-lingual Sentiment Analysis: }

Cross-lingual approaches to sentiment analysis attempt to leverage available sentiment annotations in a high-resource language for target languages which lack annotated data. This is especially important when the cost of annotating a high-quality sentiment dataset, such as the Stanford Sentiment Treebank  \cite{Socher2013b}, can be prohibitive (215,154 phrases, each annotated by 3 annotators, at 10 cents an annotation would be 64,546€!). Therefore, it is preferable to make use of those datasets that already exist. Although most approaches to cross-lingual sentiment analysis rely on Machine Translation \cite{Banea2008,Balahur2014d,Klinger2015}, this requires large amounts of parallel data,
making it less helpful for under-resourced languages.

Bilingual word embeddings have enabled cross-lingual transfer with small amounts of parallel data \cite{Artetxe2017,Lample2017} or even none at all \cite{lample2018unsupervised,Artetxe2018}, and are now used as features for state-of-the-art document-level \cite{Chen2016}, sentence-level \cite{Barnes2018b}, and targeted \cite{Hangya2018} cross-lingual sentiment analysis approaches. The objective of bilingual embeddings is to learn a shared vector space in which translation pairs have similar vector representations. This benefits under-resourced languages as a sentiment classification model trained on the source-language can be applied directly to target-language data, without the need to translate it.

\paragraph{Word Order in Sentiment Analysis: }

Pre-training sentiment classifiers with a language-modeling task represents a successful transfer learning method. \namecite{Peters2018} learn
to create contextualized embeddings by training a character-level convolutional
network to predict the next word in a sequence. Similarly, \namecite{Howard2018} introduce techniques that improve the fine-tuning of the base language-model. Likewise, 
\namecite{Devlin2018} introduce a self-attention network and adjust the language
modeling task to a cloze task, where they predict missing words in a sentence, rather than the next word given a sequence. They then fine-tune their models on downstream tasks. These models that explicitly learn word order have led to state-of-the-art results on monolingual sentiment tasks.

\paragraph{Word Reordering: }

Rule-based pre-reordering has a long tradition in Machine Translation (see \namecite{Bisazza2016} for a survey), where word order directly affects the quality of the final result. Reordering rules can be determined manually \cite{Collins2005,Gojun2012} or with data-driven approaches that either learn POS-tag based \cite{Crego2006,Crego2006b} or tree-based \cite{Neubig2012,Nakagawa2015} reordering rules. The advantage of POS-tag based rules is that they are simple to implement and do not require full parsing of the target-language sentences.

\section{Methodology}

\subsection{Corpora and Datasets}

\begin{table}[tb]
\centering
\begin{tabular}{lrrrr}
\toprule
    & & \multicolumn{1}{c}{EN} & \multicolumn{1}{c}{ES} & \multicolumn{1}{c}{CA} \\
\cmidrule(rl){2-2}\cmidrule(l){3-3}\cmidrule(l){4-4}\cmidrule(l){5-5}
 \multirow{2}{*}{\rt{Binary}}
 &$+$   & 1258 & 1216 & 682     \\
 &$-$   & 473 & 256 & 467   \\
\cmidrule(rl){2-2}\cmidrule(l){3-3}\cmidrule(l){4-4}\cmidrule(l){5-5}
 \multirow{4}{*}{\rt{4-class}}
 &$++$   & 379 & 370  & 256  \\
 &$+$    & 879 & 846  & 426   \\
 &$-$    & 399 & 218  & 409    \\
 &$--$   &  74 & 38   & 58     \\
 \cmidrule(rl){2-2}\cmidrule(l){3-3}\cmidrule(l){4-4}\cmidrule(l){5-5}
 &\textit{Total}     & 1731  & 1472     & 1149       \\
\bottomrule
\end{tabular}
\caption{Statistics for the OpeNER English (EN) and Spanish (ES) 
as well as the MultiBooked Catalan (CA) sentence-level datasets \cite{Agerri2013,Barnes2018a}}
\label{datasetstats}
\end{table}

\begin{table*}[]
\centering\scriptsize
\newcommand{\lex}[1]{{\setlength{\fboxsep}{1pt}\colorbox{lightred}{\textbf{#1}}}}
\newcommand{\nonlex}[1]{{\setlength{\fboxsep}{1pt}\colorbox{lighterred}{\textit{#1}}}}
\begin{tabular}{ll}
\toprule
\emph{\original} & Único punto \lex{negativo} el \lex{ruido} que las ventanas de madera tan típicas de la zona \nonlex{no} \nonlex{consiguen} \lex{aislar} \\[3pt]
\emph{\mtreordered} & Único \lex{negativo} punto el \lex{ruido} que las ventanas de tan típicas madera de la \nonlex{no} zona \nonlex{consiguen} \lex{aislar} \\[3pt]
\emph{\nadj} & Único \lex{negativo} punto el \lex{ruido} que las ventanas de madera tan típicas de la zona \nonlex{no} \nonlex{consiguen} \lex{aislar} \\[3pt]
\emph{\random} & aislar madera \nonlex{consiguen} típicas de el de zona las ventanas punto \lex{negativo} Único la \nonlex{no} \lex{ruido} tan que\\[3pt]
\emph{\onlylex} & UNK \lex{negativo} UNK UNK \lex{ruido} UNK UNK UNK UNK UNK UNK UNK UNK UNK UNK  \lex{aislar}\\[3pt]
\emph{\nolex} & Único punto UNK el UNK que las ventanas de madera tan típicas de la zona \nonlex{no} \nonlex{consiguen} UNK \\[3pt]
\cmidrule(lr){0-1}
\emph{Translation} & The only \lex{negative} point the \lex{noise} that the typical wooden windows in the area \nonlex{don't manage} to \lex{block} \\[3pt]
\bottomrule
\end{tabular}
\caption{An example of a negative Spanish sentence (\original) with the five reordering
transformations applied, as well as its English translation. The \lex{bold tokens} are words found in the sentiment lexicon, and the \nonlex{italic words} are words that convey sentiment in this instance, but are not in the lexicon}
\label{example}

\end{table*}

At document-level, bag-of-words models are often expressive enough to give good results without relying on word order \cite{Meng2012,Iyyer2015}. But because we are interested in word-order effects in cross-lingual sentiment analysis, we require datasets that are annotated at a fine-grained level, \ie sentence- or aspect-level.

For this reason, we use the English and Spanish OpeNER corpora of hotel reviews \cite{Agerri2013} as well as the Catalan MultiBooked Dataset \cite{Barnes2018a}. Statistics on the corpora are shown in Table \ref{datasetstats}. Each sentence is annotated for four classes of sentiment (strong positive, positive, negative, and strong negative). We use the English subset for training our classifiers and the Spanish and Catalan for testing the effects of word order on the target languages. Although these datasets are relatively small, they are all annotated similarly and
are in-domain, which avoids problems with mapping labels or domain shifts.

\subsection{Bilingual Word Embeddings}

VecMap \cite{Artetxe2016,Artetxe2017} creates bilingual embeddings by learning an orthogonal projection between two pre-computed monolingual vector spaces and requires only a small bilingual dictionary. We use the publicly available GoogleNews vectors for the English (available at \url{https://code.google.com/archive/p/word2vec/}), and for Spanish and Catalan we create skip-gram embeddings with 300 dimensions trained on Wikipedia data. The bilingual dictionaries are translated sentiment lexicons \cite{HuandLiu2004} with 5700 pairs for English -- Spanish (5271 for English -- Catalan).

\subsection{Experimental Setup}

In order to test whether a sentiment classifier trained on bilingual embeddings is sensitive to word order, we test classifiers on six versions of the target-language sentiment data, which we describe in the following section. An example of these six versions is shown in Table \ref{example}.

\paragraph{\original: } We test the model on the original data with no changes in word order.

\paragraph{\mtreordered: }A competing hypothesis is that a full pre-reordering of the target-side sequences will be more familiar to the sentiment classifier trained on English and therefore lead to better results. We implement POS-tag based rewrite rules \cite{Crego2006,Crego2006b}, which are then applied to the target-language test data before testing.

\paragraph{\nadj: }Given that adjectives are important for sentiment analysis, we hypothesize that adjusting the order of nouns and adjectives should be beneficial if the classifier is learning source-language word order. Therefore, we implement a simple reordering which places Spanish and Catalan adjectives before, rather than after, the noun they modify.

\paragraph{\random: }We randomly permutate the order of the target-language sentences. If the sentiment classification models take the target language word order into consideration, this should lead to poor results.

\paragraph{\onlylex and \nolex: }Finally, we provide two baselines for clarification. The \onlylex experiment removes all words which do
not appear in the Hu \& Liu sentiment lexicon \cite{HuandLiu2004}. If our systems
take word order into account, they should be affected negatively by this, as the
resulting sentence does not resemble the normal word order. If, however, the models
are relying on keywords, this will have little effect.

For the \emph{\nolex} experiment, we remove all of the words
in a phrase which appear in the sentiment lexicon. If the models are attending to
sentiment keywords, this approach should lead to the worst performance. 

\paragraph{Baselines: }We perform additional experiments with monolingual and Machine Translation (MT)-based cross-lingual approaches. For the former, we use the Google API (available at \url{https://translate.google.com/}) and translate the target-language data to English. 

For both baseline setups, we only test the \random reordering, \onlylex, and \nolex approaches. Additionally, the monolingual setup is not comparable to the MT and cross-lingual versions, as we must use the target-language data for training, development, and testing (70\%/10\%/20\%).

\begin{table*}[th!]
\newcommand{\sep}{\cmidrule(r){4-6}\cmidrule(r){7-9}}
\newcommand{\sepp}{\cmidrule(r){4-4}\cmidrule(r){5-5}\cmidrule(r){6-6}\cmidrule(r){7-7}\cmidrule(r){8-8}\cmidrule(r){9-9}}

\definecolor{green}{RGB}{150,255,150}
\definecolor{blue}{RGB}{150,150,255}
\definecolor{pink}{RGB}{255,182,193}

\newcommand{\bestproj}[1]{{\setlength{\fboxsep}{0pt}\colorbox{lightblue}{\textbf{#1}}}}
\newcommand{\bestmono}[1]{{\setlength{\fboxsep}{0pt}\colorbox{lightgreen}{\textbf{#1}}}}
\newcommand{\bestmt}[1]{{\setlength{\fboxsep}{0pt}\colorbox{pink}{\textbf{#1}}}}

\setlength\tabcolsep{4pt}
\renewcommand*{\arraystretch}{0.5}
\centering
\begin{tabular}{lllcccccccccccc}
\toprule
&& & \multicolumn{3}{c}{4-class} & \multicolumn{3}{c}{Binary} \\
\sep
\multirow{20}{*}{\rt{BWE}} 
	    &&& \bilstm & \cnn & SVM & \bilstm & \cnn & SVM \\
	    \cmidrule(r){4-4}\cmidrule(r){5-5}\cmidrule(r){6-6}\cmidrule(r){7-7}\cmidrule(r){8-8}\cmidrule(r){9-9}
	& \multirow{6}{*}{\rt{EN-ES}}
		& \original 	 & 33.3 \scriptsize{(1.8)}   & 35.4 \scriptsize{(1.1)} & 34.9 & 64.9 \scriptsize{(0.9)}  & 60.0 \scriptsize{(1.4)} & 66.6\\ 
		&& \mtreordered  & \bestproj{34.0 \scriptsize{(1.6)}} & 35.6 \scriptsize{(1.4)} & 34.9 & \bestproj{65.1 \scriptsize{(1.3)}}  & 60.1 \scriptsize{(1.3)} & 66.6\\ 
		&& N-ADJ         & \bestproj{34.0 \scriptsize{(1.8)}} & \bestproj{35.8 \scriptsize{(1.2)}} & 34.9 & 65.0 \scriptsize{(1.2)}  & \bestproj{60.2 \scriptsize{(1.4)}} & 66.6\\ 
		&& \random       & 33.2 \scriptsize{(1.3)} & 35.3 \scriptsize{(1.1)} & 34.9 & 63.9 \scriptsize{(2.3)}  & 58.8 \scriptsize{(0.9)} & 66.6\\ 
		&& \onlylex      & 28.2 \scriptsize{(3.8)} & 26.9 \scriptsize{(2.5)} & 30.7 & 57.6 \scriptsize{(5.5)}  & 34.2 \scriptsize{(5.5)} & 53.0\\ 
		&& \nolex        & 31.9 \scriptsize{(1.6)} & 33.2 \scriptsize{(1.4)} & 33.3 & 61.1   & 57.1 \scriptsize{(2.8)} & 63.4\\ 
	\sepp
	& \multirow{6}{*}{\rt{EN-CA}}
  		& \original      &  37.0 \scriptsize{(1.4)} & 37.4 \scriptsize{(1.5)} & 33.2 & 64.0 \scriptsize{(1.1)} & 61.9 \scriptsize{(6.8)} & 68.2\\ 
		&& \mtreordered  &  \bestproj{37.8 \scriptsize{(1.2)}} & 37.9 \scriptsize{(1.5)} & 33.2 & \bestproj{65.6 \scriptsize{(1.5)}} & 62.6 \scriptsize{(5.8)} & 68.2\\ 
		&& N-ADJ         &  37.7 \scriptsize{(1.5)} & \bestproj{38.1 \scriptsize{(1.6)}} & 33.2 & 65.5 \scriptsize{(1.5)} & \bestproj{62.8 \scriptsize{(6.3)}} &  68.2\\ 
		&& \random       &  35.7 \scriptsize{(1.0)} & 35.6 \scriptsize{(1.5)} & 33.2 & 63.3 \scriptsize{(0.8)} & 60.8 \scriptsize{(5.5)} & 68.2\\ 
		&& \onlylex      &  28.2 \scriptsize{(1.8)} & 25.7 \scriptsize{(3.2)} & 23.8 & 49.9 \scriptsize{(4.3)} & 40.5 \scriptsize{(6.7)} & 39.1 \\ 
		&& \nolex        &  35.9 \scriptsize{(1.7)} & 34.3 \scriptsize{(1.8)} & 31.2 & 61.7 \scriptsize{(1.1)} & 58.1 \scriptsize{(5.3)} & 63.1\\

\\
\hline \\

\multirow{12}{*}{\rt{MT}}
	& \multirow{4}{*}{\rt{EN-ES}}
  		& \original      & \bestmt{46.5 \scriptsize{(1.2)}} & \bestmt{41.2 \scriptsize{(3.7)}} & 44.6 & \bestmt{71.8 \scriptsize{(1.1)}} & \bestmt{64.3 \scriptsize{(1.6)}} & 70.7\\ 
		&& \random       & 46.0 \scriptsize{(1.8)} & 38.9 \scriptsize{(3.9)} & 44.6 & 71.0 \scriptsize{(1.4)} & 62.2 \scriptsize{(1.5)} & 70.7\\ 
		&& \onlylex      & 32.9 \scriptsize{(2.5)} & 28.2 \scriptsize{(4.4)} & 36.2 & 63.0 \scriptsize{(3.8)} & 44.6 \scriptsize{(3.5)} & 51.9\\ 
		&& \nolex        & 41.8 \scriptsize{(0.7)} & 37.0 \scriptsize{(3.0)} & 41.6 & 63.0 \scriptsize{(1.1)} & 54.8 \scriptsize{(2.6)} & 66.2\\ 
	\sepp
	& \multirow{4}{*}{\rt{EN-CA}}
  		& \original      & \bestmt{51.5 \scriptsize{(3.1)}} & \bestmt{44.1 \scriptsize{(4.3)}} & 46.8 & \bestmt{79.9 \scriptsize{(1.5)}} & \bestmt{72.8 \scriptsize{(2.0)}}  & 74.2  \\ 
		&& \random       & 49.7 \scriptsize{(1.4)} & 37.7 \scriptsize{(3.6)} & 46.8 & 76.5 \scriptsize{(2.2)} & 66.4 \scriptsize{(1.4)}  & 74.2  \\ 
		&& \onlylex      & 32.0 \scriptsize{(2.7)} & 32.5 \scriptsize{(4.0)} & 36.1 & 58.4 \scriptsize{(7.8)} & 57.5 \scriptsize{(2.9)}  & 43.7  \\ 
		&& \nolex        & 48.4 \scriptsize{(2.0)} & 40.9 \scriptsize{(2.9)} & 46.2 & 75.6 \scriptsize{(1.6)} & 65.6 \scriptsize{(2.8)}  & 70.4  \\[3pt]

\bottomrule
\end{tabular}
\caption{Macro \F results for all corpora and techniques. We denote
  the best performing bilingual embedding
  method per column with a \bestproj{blue box}, the best MT method with a \bestmt{pink box}. We do not denote bag-of-words SVM results, as they are invariant to word order}
\label{results:all}
\end{table*}

\begin{table*}[th!]
\newcommand{\sep}{\cmidrule(r){4-6}\cmidrule(r){7-9}}
\newcommand{\sepp}{\cmidrule(r){4-4}\cmidrule(r){5-5}\cmidrule(r){6-6}\cmidrule(r){7-7}\cmidrule(r){8-8}\cmidrule(r){9-9}}

\definecolor{green}{RGB}{150,255,150}
\definecolor{blue}{RGB}{150,150,255}
\definecolor{pink}{RGB}{255,182,193}

\newcommand{\bestproj}[1]{{\setlength{\fboxsep}{0pt}\colorbox{lightblue}{\textbf{#1}}}}
\newcommand{\bestmono}[1]{{\setlength{\fboxsep}{0pt}\colorbox{lightgreen}{\textbf{#1}}}}
\newcommand{\bestmt}[1]{{\setlength{\fboxsep}{0pt}\colorbox{pink}{\textbf{#1}}}}

\setlength\tabcolsep{4pt}
\renewcommand*{\arraystretch}{0.5}
\centering
\begin{tabular}{lllcccccccccccc}
\toprule
&& & \multicolumn{3}{c}{4-class} & \multicolumn{3}{c}{Binary} \\
\sep
\multirow{13}{*}{\rt{Monolingual}}
    & \multirow{4}{*}{\rt{ES}}
  		& \original  & \bestmono{43.2 \scriptsize{(3.3)}} & \bestmono{36.2 \scriptsize{(2.2)}} & 32.1  & \bestmono{68.5 \scriptsize{(3.4)}} & \bestmono{64.8 \scriptsize{(2.3)}} & 52.7 \\
		&& \random 	 & 42.5 \scriptsize{(2.6)} & 32.7 \scriptsize{(1.8)} & 32.1  & 67.5 \scriptsize{(4.2)} & 63.1 \scriptsize{(2.7)} & 52.7 \\ 
		&& \onlylex  & 27.0 \scriptsize{(0.5)} & 21.2 \scriptsize{(4.5)} & 27.0  & 45.2 \scriptsize{(0.0)} & 47.9 \scriptsize{(3.9)} & 45.2 \\ 
		&& \nolex 	 & 37.9 \scriptsize{(1.9)} & 34.3 \scriptsize{(2.0)} & 30.3  & 64.7 \scriptsize{(2.7)} & 65.0 \scriptsize{(0.9)} & 51.8 \\ 
		\sepp
	& \multirow{4}{*}{\rt{CA}}
  		& \original  & \bestmono{48.6 \scriptsize{(1.6)}} & \bestmono{46.2 \scriptsize{(0.8)}}  & 46.8  & \bestmono{77.1 \scriptsize{(1.3)}} & \bestmono{76.4 \scriptsize{(1.2)}} & 75.0  \\
		&& \random   & 47.4 \scriptsize{(1.9)} & 43.9 \scriptsize{(3.0)}  & 46.8  & 73.6 \scriptsize{(1.3)} & 71.9 \scriptsize{(1.9)} & 75.0  \\ 
		&& \onlylex  & 20.3 \scriptsize{(2.8)} & 27.4 \scriptsize{(3.2)}  & 16.7  & 40.1 \scriptsize{(1.5)} & 56.4 \scriptsize{(5.2)} & 39.6  \\ 
		&& \nolex    & 47.5 \scriptsize{(0.6)} & 45.8 \scriptsize{(1.6)}  & 45.8  & 75.0 \scriptsize{(1.6)} & 74.5 \scriptsize{(1.1)} & 74.8  \\ 
\bottomrule
\end{tabular}
\caption{Macro \F results for all  monolingual models. Although these results are not comparable to BWE or MT, as they are calculated on a smaller dataset, we provide them as a general reference to what results can be expected under monolingual conditions. We denote the best monolingual method
  with a \bestmono{green box}}
\label{results:mono}
\end{table*}

\subsection{Models}

To test our hypotheses, we compare three different classifiers: a Support Vector Machine (SVM) with Bag-of-Embeddings feature representations, a Convolutional Neural Network (\cnn) \cite{Santos2014,Severyn2015}, and a Bidirectional Long Short-Term Memory Network (\bilstm) \cite{Luong2015}. Each of these classifiers theoretically has an increasing reliance on word order. Note that we do not use the bilingual sentiment model \cite{Barnes2018b}, as it jointly learns both projection and classifier and cannot be used as input to the \cnn and \bilstm. Although the SVM does not take into account word order at all, it is a strong baseline for sentiment analysis \cite{Kiritchenko2014c}. The \cnn considers only local word order, while the \bilstm relies on both local and long-distance dependencies. 

For the neural models, we train five classifiers on five random seeds and report the mean and standard deviation of the macro \F score, while we only report the macro \F score of a single run for the SVM.

\paragraph{\bilstm} We implement a single-layered \bilstm classifier with a 100-dimensional hidden layer, which passes the concatenation of the two final hidden states to a softmax layer for classification. The cross-lingual model is initialized with the pre-trained bilingual embeddings (monolingual embeddings for the monolingual and translation models), use dropout of $0.3$ for regularization, and are trained for 30 epochs with a batch size of 32 using Adam as an optimizer. We choose the parameter for training epochs on the source-language development set and test this model on the target-language data.

\paragraph{\cnn} The \cnn has a single convolutional layer with filters $\in \{3,4,5\}$ followed by a max-pooling layer of length $2$. The pooled representation of the sentence is passed to a feed-forward layer and finally a softmax layer of size $\R^{|L|}$ where $L$ is the set of labels. The optimization is the same as the \bilstm, with dropout applied after the feed-forward layer.

\paragraph{SVM}
 Finally, we implement a baseline bag-of-embeddings SVM. For each sentence in the dataset, we create an averaged embedding representation $A= \frac{1}{n} (\sum_{i=1}^{n} e(t_{i}))$ where $e(t_{i})$ is the embedding representation of the $i$th token in the sentence $S \in \{t_{1}, t_{2}, \ldots, t_{n}\}$ of length $n$. For the cross-lingual approaches we use the bilingual embeddings (monolingual embeddings for the monolingual and translation approaches) and tune the $c$ parameter on the source-language development set.

\begin{table*}[]
\centering
\newcommand{\sepp}{\cmidrule(rl){1-1}\cmidrule(rl){2-2}\cmidrule(rl){3-3}}
\begin{tabular}{lll}
\toprule
model & text & prediction \\
\sepp
\emph{\original} & relación calidad precio muy buena & negative \\
\emph{\mtreordered} & relación muy buena calidad precio & positive \\
\textit{translation} & very good quality price relationship & positive \\
\sepp
\emph{\original} & hotel perfecto & negative \\
\emph{\mtreordered} & perfecto hotel & positive \\
\textit{translation} & perfect hotel & positive \\
\sepp
\emph{\original} & el desayuno muy bueno . & negative \\
\emph{\mtreordered} & el muy bueno desayuno . & positive \\
\textit{translation} & the breakfeast (was) very good & positive \\
\sepp
\emph{\original} & gestión nefasta . & positive \\
\emph{\mtreordered} & nefasta gestión . & negative \\
\textit{translation} & terrible management & negative \\
\bottomrule
\end{tabular}
\caption{Examples where reordering improves results over original on binary English-Spanish setup with the BiLSTM classifier}
\label{helpful_examples}

\end{table*}

\section{Results}

Table \ref{results:all} shows the results of all experiments. Firstly, reordering the test data improves the results on all of the eight experiments (we do not consider SVM experiments to calculate improvements as they are invariant to word order). Specifically, the \mtreordered approach improves the \bilstm results the most on all experiments, while the simpler \nadj flip is the best performing setup with \cnns. This indicates that local word reordering has more of an effect on \cnns, while the global reordering can be more helpful to \bilstms. While the improvements from reordering are often small (0.2 - 1.6 percentage points (ppt)), they are stable. 

While it is the case that in both of the target languages \nadj combinations can have a different meaning if the order is switched (for example ``el amigo viejo" and ``el viejo amigo"), the practical relevance of these order changes is minimal: in the Spanish dataset, of 978 occurrences of \nadj, only 23 (2.35\%) occur as well with a \adjn order; in the Catalan dataset, of 745 occurrences of \nadj, only 8 (1.07\%) occur as well with a \adjn order.

\random has a more substantial negative effect on monolingual models (an average decrease of 1.6 ppt for \bilstm and 3.0 ppt for \cnn) and MT-based models (1.6/4.3 ppt, respectively) than bilingual embedding models (0.8/1.1). This indicates that noise from the embedding projection renders it more difficult for models to use word order in the cross-lingual setup.

Additionally, \random has a larger effect on the \cnn (an average loss of 1.1 ppt) than on the \bilstm (0.8). This is likely because the \cnn relies on specific combinations of n-grams in order to correctly classify a sentence. If these are not present, the filters are not effective at classification.

Although they are not comparable (the test datasets have fewer examples), the monolingual models generally perform better than the cross-lingual versions, except for the SVM classifiers. The machine translation approaches perform better than the cross-lingual embedding methods.

The classification models display different trends across the setups. On the monolingual and machine translations setups, the \bilstm is the strongest model, followed by the \cnn and SVM (SVM and \cnn, respectively for machine translation). With bilingual embeddings, however, the SVM outperforms both the \bilstm and \cnn on the Spanish binary setup, while the \cnn is strongest on the multiclass.  This shows that \bilstm displays a different behavior with bilingual embeddings.

The machine translation models perform well and surprisingly suffer less than monolingual models (an average decrease of 15.4 ppt for MT \bilstm and \cnn models vs. 20.6 for monolingual)  from using only features from the sentiment lexicon (\onlylex). This suggests that MT models rely more on these keywords while ignoring word order effects to a higher degree.

Finally, the \nolex and \onlylex baselines perform poorly, with \onlylex often more than 20 ppt below the performance of \original. This is due largely to the low coverage of the sentiment lexicon used in this work, as many full sentences were completely unked (38\% for Spanish, 43\% for Catalan). This also explains the similar performances of \original and \nolex.

\section{Analysis}

Reordering tends to help both the \bilstm and \cnn models with shorter examples (less than eight tokens long) where adjective order can easily be changed to resemble English word order, such as the examples shown in Table \ref{helpful_examples}. In longer instances (more than ten tokens), however, the reordering either introduces too much noise or does not affect the final prediction. The current reordering models are therefore more adequate for sentiment tasks that deal with shorter texts, such as aspect- or sentence-level, rather than document-level sentiment analysis.

\section{Conclusion and Future Work}

In this work, we have shown that neural networks that rely on bilingual embeddings as features
are sensitive to differences in source- and target-language word order and subsequently benefit from reordering the target language test data. The gains, however, are still relatively small, which suggests that currently bilingual embeddings introduce too much noise for a classifier to generalize well to the target language.

Although our reordering approach does improve the neural models, these more expressive models are still outperformed by the linear SVM with bag-of-embeddings representations. This is likely a side effect of the noise introduced by the bilingual embeddings. At test time, the model receives as input embeddings that are \emph{similar} but not necessarily the same as at training. In the future, it may be helpful to develop models which are more robust to this noise, or alternatively to use low-resource machine translation techniques \cite{artetxe2018iclr,lample2018unsupervised,artetxe2018emnlp,lample-etal-2018-phrase}

Given that language modeling pre-training is beneficial for state-of-the-art results in monolingual sentiment analysis, it is important to realize that cross-lingual models based on bilingual word embeddings do not currently benefit from word order learned in the source language. In the future, we would like to pre-train bilingual language models for cross-lingual sentiment analysis.

\bibliographystyle{fullname}
\bibliography{lit}

\end{document}